\pdfoutput=1
\documentclass[11pt]{article}

\usepackage[preprint]{acl}

\usepackage{times}
\usepackage{latexsym}

\usepackage[T1]{fontenc}
\usepackage[utf8]{inputenc}

\usepackage{microtype}

\usepackage{inconsolata}

\usepackage{graphicx}

\usepackage{enumitem}
\usepackage{xcolor}
\usepackage{amsthm}
\usepackage{booktabs}
\usepackage{subcaption}
\usepackage{multirow}
\usepackage{mdframed}
\usepackage{multirow}

\newtheorem{definition}{Definition}[section]
\newcommand{\equalcontrib}{\textsuperscript{\textdagger}}  % 将共同作者标志定义为星号
\newcommand{\correspondauthor}{\textsuperscript{*}}

\title{Advancing and Benchmarking Personalized Tool Invocation for LLMs}

\author{
    \textbf{Xu Huang}\textsuperscript{1}\equalcontrib,
    \textbf{Yuefeng Huang}\textsuperscript{1}\equalcontrib,
    \textbf{Weiwen Liu}\textsuperscript{2}\correspondauthor,
    \textbf{Xingshan Zeng}\textsuperscript{3},
    \textbf{Yasheng Wang}\textsuperscript{3},\\
    \textbf{Ruiming Tang}\textsuperscript{3},
    \textbf{Hong Xie}\textsuperscript{1},
    \textbf{Defu Lian}\textsuperscript{1}\correspondauthor
    \\
    \\
    \textsuperscript{1}University of Science and Technology of China,
    \textsuperscript{2}Shanghai Jiao Tong University,
    \textsuperscript{3}Huawei Noah's Ark Lab
    \\
    \small{
     \href{mailto:xuhuangcs@mail.ustc.edu.cn}{xuhuangcs@mail.ustc.edu.cn}
     \href{mailto:huang\_yuefeng@mail.ustc.edu.cn}{huang\_yuefeng@mail.ustc.edu.cn}
     }
}

\begin{document}
\maketitle
\begin{abstract}
% This document is a supplement to the general instructions for *ACL authors. It contains instructions for using the \LaTeX{} style files for ACL conferences.
% The document itself conforms to its own specifications, and is therefore an example of what your manuscript should look like.
% These instructions should be used both for papers submitted for review and for final versions of accepted papers.

% Tool invocation 是Large Language Models （LLMs）扩展自身能力的一个重要手段，近期受到了广泛的关注。其支持LLMs可以通过一系列工具调用解决复杂问题，并可以获取最新世界知识。然而，现有的工作仅仅考虑到LLMs的基础工具调用能力，即能否调用工具解决问题，但没有考虑到在个性化约束下的工具调用问题。我们在本工作中首次提出了Personalized Tool Invocation的概念，并定义了两个任务：Tool Preference和Imcomplete Query。Tool Preference中考虑了用户在面临多个功能相似的工具下的偏好性问题；Imcomplete Query考虑了用户在query中会缺失部分工具参数，要求模型从用户profile中识别并提取。我们针对这两个主要任务设计了一套数据合成框架 PersonalTool，并构造了第一个个性化工具调用的Benchmark：PTBench。我们利用合成的数据对一系列开源模型进行了微调实验，验证了所提出框架的有效性并给出了一些insights

Tool invocation is a crucial mechanism for extending the capabilities of Large Language Models (LLMs) and has recently garnered significant attention. It enables LLMs to solve complex problems through tool calls while accessing up-to-date world knowledge. However, existing work primarily focuses on the fundamental ability of LLMs to invoke tools for problem-solving, without considering personalized constraints in tool invocation.
In this work, we introduce the concept of Personalized Tool Invocation and define two key tasks: Tool Preference and Profile-dependent Query. Tool Preference addresses user preferences when selecting among functionally similar tools, while Profile-dependent Query considers cases where a user query lacks certain tool parameters, requiring the model to infer them from the user profile.
To tackle these challenges, we propose \textbf{PTool}, a data synthesis framework designed for personalized tool invocation. Additionally, we construct \textbf{PTBench}, the first benchmark for evaluating personalized tool invocation. We then fine-tune various open-source models, demonstrating the effectiveness of our framework and providing valuable insights. Our benchmark is public at \url{https://github.com/hyfshadow/PTBench}.

\end{abstract}

{
\renewcommand{\thefootnote}{\textdagger}
\footnotetext[1]{Equal Contributions.}
\renewcommand{\thefootnote}{*}
\footnotetext[2]{Corresponding: \href{mailto:wwliu@sjtu.edu.cn}{wwliu@sjtu.edu.cn},\href{mailto:liandefu@ustc.edu.cn}{liandefu@ustc.edu.cn}}
\renewcommand{\thefootnote}{\fnsymbol{footnote}}
}

\section{Introduction}

Recently, large language models (LLMs) have demonstrated remarkable capabilities in natural language processing tasks, particularly in human-computer interaction, where they can effectively comprehend user queries and provide reasonable responses~\cite{zhao2023survey}. However, the knowledge embedded within LLMs is not inherently up-to-date, as updating these models requires extensive retraining with large-scale data, which incurs significant time and economic costs. To equip LLMs with the ability to solve complex problems and access the latest information, tool invocation capabilities are essential. For instance, LLMs can leverage mathematical tools to decompose and solve intricate mathematical problems or utilize internet APIs~\cite{liu2025toolace,qin2024toolllm} and search engines~\cite{schick2024toolformer, nakano2021webgpt} to retrieve the most recent knowledge.

\begin{figure}[t]
    \centering
    \includegraphics[width=\linewidth]{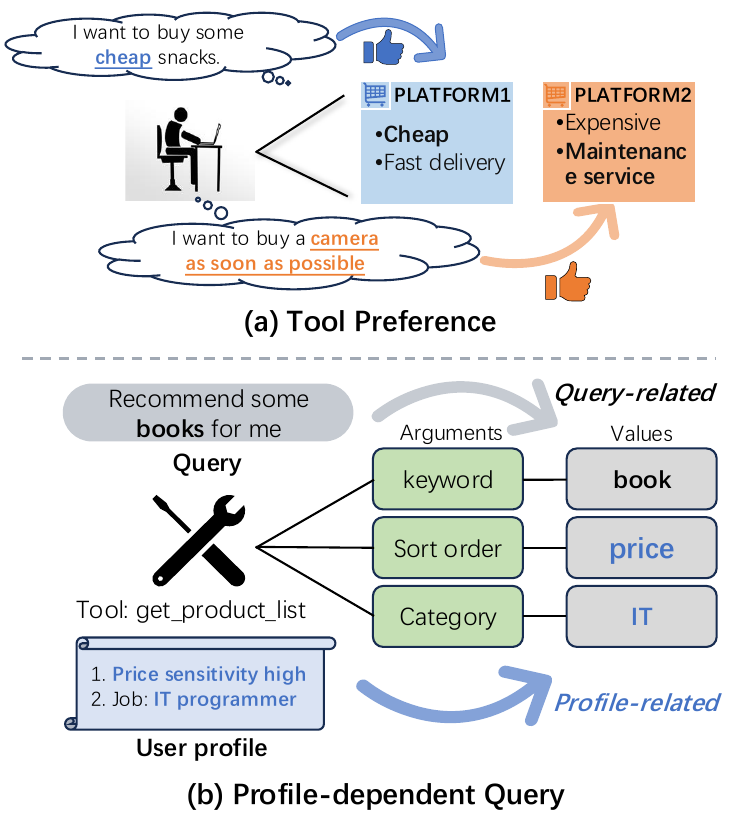}
    \caption{Example of Personalized Tool Invocation. (a) Tool Preference: Users may prefer different tools for similar functionalities depending on the query context. (b) Profile-dependent Query: Certain tool parameters may be missing from the user's query and need to be inferred from the user profile.}
    \label{fig:enter-label}
\end{figure}

Existing research on enhancing LLMs's tool invocation abilities primarily focuses on improving fundamental capabilities~\cite{qin2024toolllm,berkeley-function-calling-leaderboard,lin2024hammer}, such as ensuring adherence to the required tool invocation syntax, comprehending tool functionalities, interpreting explicit user instructions, and extracting tool parameters. However, in real-world applications, user intents are often implicit rather than explicitly stated, requiring models to infer based on user-personalized profiles and behavioral history before invoking the appropriate tools. Two common scenarios illustrate this challenge on personalized tool invocation: 
\textbf{(1) Tool Preference}. When multiple tools offer similar functionalities, users often exhibit specific preferences. For example, in online shopping, users may choose different platforms depending on their preferences for particular product categories. Some users may prioritize platforms with superior maintenance services when purchasing high-value electronic products, despite the higher cost, while preferring platforms with faster delivery when buying inexpensive daily necessities. Inferring such preferences necessitates reasoning from user attributes, such as age, interests, and purchasing behavior.
\textbf{(2) Profile-dependent Query}. In everyday scenarios, users tend to express their needs concisely and omit crucial details. For instance, a user might simply request, "Order me a hamburger from KFC," without specifying essential information such as the delivery address, recipient contact details, or preferred delivery time. This requires the model to infer the missing information based on the user profile, such as the user's work location, current time, and contact information, ensuring a seamless and accurate tool invocation process.

In this work, we propose the novel task of personalized tool invocation, aiming to address the aforementioned critical challenges. To enhance and systematically evaluate a model's ability in personalized tool invocation, we further introduce an automated data synthesis framework for this task, termed as \textbf{PTool}, which consists of three key stages: tool generation, user profile construction, and user behavior simulation.
Firstly, we consider multiple commonly used real-world scenarios, where each scenario contains multiple functionally similar platforms organized in a hierarchical tree structure. We then leverage an advanced large language model (LLM) to recursively decompose platform functionalities using a depth-first expansion approach, progressively refining them until distinct tool APIs are defined for each functional category. This ensures that the generated tools comprehensively cover the functional demands of the given scenarios, thereby increasing the diversity of tools.
Secondly, we abstract and summarize platform features and API parameters to extract both basic user attributes and personalized characteristics, including psychological traits and behavioral tendencies. To construct a diverse set of user profiles, we employ a bottom-up clustering approach for feature induction and a top-down assignment strategy for attribute allocation.
Finally, we exploit the role-playing capabilities of LLMs to simulate user behaviors based on the assigned user profiles, generating both historical interactions and potential user queries. To establish reliable ground-truth labels, we further integrate a multi-agent framework that conditions query generation on user profiles. Following manual review and annotation, we construct \textbf{P}ersonalized \textbf{T}ool\textbf{Bench} (\textbf{PTBench}), the first benchmark designed to evaluate large models' ability in personalized tool invocation, consisting of 1,083 high-quality annotated data samples. Our key contributions are summarized as follows:
\begin{itemize}[leftmargin=*]
    \item We propose the first paradigm for personalized tool invocation, incorporating both user tool preferences and profile-dependent user queries, two key challenges in real-world applications.
    \item We develop a systematic personalized data synthesis framework and construct PTBench, the first benchmark for personalized tool invocation, enabling a comprehensive evaluation of models' ability to invoke tools based on user information.
    \item We demonstrate that training open-source models on our synthesized dataset significantly improves personalized tool invocation capabilities, while also enhancing general tool invocation without compromising other general abilities.
\end{itemize}

\section{Related Work}

\subsection{Tool Invocation}
Tool invocation (also termed tool calling) involves tool selection from candidate tools and parameter extraction from queries. Existing works can be categorized into two tuning-free and tuning-based methods~\cite{qu2025tool,liu2023controlllm}. Tuning-free methods mainly rely on the prompt strategy with few-shot learning, involving encouraging LLM to reason by providing examples~\cite{yao2022react}, rewriting tool documentation with LLMs to enhance the comprehension~\cite{yuan2024easytool}, summarizing tool description with more concise and precise sentence~\cite{xu-etal-2024-concise}, leveraging multi-agent collaboration to decompose the tool-calling task~\cite{shi-etal-2024-learning}. Tuning-based methods leverage tool-learning samples to train existing LLMs, where the research problems comprise data collection and training strategy. Toolformer~\cite{schick2024toolformer} and ToolkenGPT~\cite{hao2024toolkengpt} add a special tool-related token into the vocabulary, switching the decoding process into tool selection and calling. Some works leverage advanced LLM to synthesize tool-calling samples to improve the tool-invocation ability of lightweight models, demonstrating the efficiency of the distillation from advanced models~\cite{qin2024toolllm,yang2023gpttools,liu2025toolace}.

\subsection{Personalized LLMs}

Personalized LLMs represent LLMs that have been adapted to align with user preferences and characteristics~\cite{zhang2024personalization}. Existing works mainly focus on the generation of personalized texts or applications in information systems. LLMs are customized as personal conversational AI assistants for various domains, including education~\cite{kasneci2023chatgpt,dan2023educhat,park2024empowering}, healthcare~\cite{belyaeva2023multimodal,abbasian2024knowledge,jin2024health}, finance~\cite{liu2023fingpt,lakkaraju2023can}, legal~\cite{nguyen2023brief}, and etc. User profiles are provided via prompts or hidden representation, leading the model to generate personalized text in the dialog. Personalized LLMs have been extensively applied in information systems such as recommender systems~\cite{llm4recsurvey,chen2024large}. LLMs are leveraged as an augmentation module for traditional recommender systems, serving as the content interpreter~\cite{bao2023tallrec,li2023exploring,yang2023palr}, the knowledge base~\cite{xi2024towards,wei2024llmrec}, or the explainer~\cite{lei2024recexplainer,wang-etal-2023-llm4vis}. Also, many works directly deploy LLMs as the direct recommenders via prompt techniques~\cite{lyu-etal-2024-llm,hou2024large} or fine-tuning~\cite{zhang2023recommendation}.
However, there is no work considering personalization in tool learning. This work is the first to propose personalized tool invocation for LLMs.
\section{Personalized Tool Invocation}
We innovatively consider a practical and high-demand scenario in LLM tool invocation: \textbf{personalized tool invocation}. This scenario requires the model to leverage user-specific information when selecting and configuring tools to address user needs. In this chapter, we formally define the task of personalized tool invocation.  

Given an LLM with model parameters \(\theta\), the general tool invocation task requires the model, when provided with a query \(q\) and a set of candidate tools \(T\), to select the appropriate tool \(t^i\) and populate its corresponding parameters \(a^i_1, \cdots, a^i_m\), forming the solution \(A = [(t^i, a^i_1, \cdots, a^i_m), \cdots]\)

In conventional formulations of this task, correctness is typically determined by whether the selected tool successfully resolves the query. However, this setting overlooks the fact that multiple tools may serve the same function (e.g., APIs from different platforms with similar capabilities), and that users often have preferences for certain tools—a concept we refer to as \textbf{tool preference}, defined as follows:
\begin{definition}(Tool Preference) User $u$ prefers $t^1$ for query $q_1$ and $t^2$ for query $q_2$, where $q_1, q_2$ can be solved by both $t^1$ and $t^2$:
\begin{equation}
    t^1 \succ_{(u,q_1)} t^2;\quad t^2 \succ_{(u,q_2)} t^1
\end{equation}
\end{definition}

Moreover, in \( A \), both tool selection and parameter values are determined solely based on the information contained in the query. For instance, consider the query: "Book me a flight from Los Angeles to New York at 8:45 AM tomorrow". However, in real-world scenarios, users often do not provide such detailed query information. Instead, they may omit certain essential details required for tool invocation, meaning that the model cannot extract all necessary parameters from the query alone. We refer to this personalized scenario as an \textbf{profile-dependent query}, defined as follows:
\begin{definition}(Profile-dependent Query) Given the profile of the user $u$ as $P_u$,  the query $q$ and the solution $A$, there exists value $\alpha \in A$, $\alpha \in P_u$ and $\alpha \notin q$, then the query $q$ is called profile-dependent query.
\end{definition}

\section{Personalized Tool Invocation Data Synthesis}

\begin{figure*}
    \centering
    \includegraphics[width=1.0\linewidth]{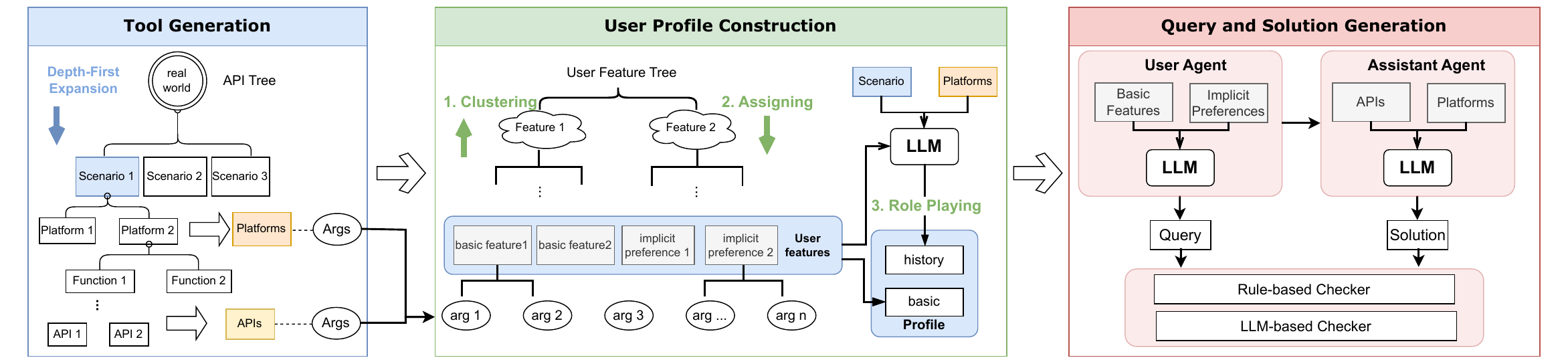}
    \caption{Framework of our personalized tool invocation data synthesis framework: PTool. The pipeline comprises three stages: Tool Generation, User Profile Generation and Query and Answer Generation.}
    \label{fig:framework}
\end{figure*}

To address the two challenges in personalized tool invocation mentioned above, we propose an automated data synthesis framework, PTool, for generating high-quality training and evaluation data for personalized tool invocation. The framework consists of three key stages: \textbf{Tool Generation}, \textbf{User Profile Construction}, and \textbf{Query and Solution Generation}, as illustrated in Figure~\ref{fig:framework}. The detailed processes of each stage are described in the subsequent parts of this section.

\subsection{Tool Generation}
% 为了构建多样化的工具库，
% 1. 为了保证工具的多样化
% 2. 如何得到同功能的工具，用于tool preference
% Tool Tree，DFS expansion，

To cover the majority of scenarios encountered in daily life, we first constructed a diversified tool library across multiple contexts. Inspired by existing work, we employed an advanced Large Language Model (LLM)-based data synthesis method to generate APIs. Similar to ToolACE, we also developed a structure akin to an API Tree, which allows for the generation of diverse tools.

Specifically, we initially define several demand scenarios from everyday life (e.g., shopping, food delivery, office) as the first-level nodes of the tree. Then, using a depth-first expansion approach, we iteratively refine the functionality at each node until we derive specific API descriptions as the leaf nodes. Notably, in order to generate data that enhances the model's Tool Preference capability, tools with similar functionalities are required. However, this API Tree expansion approach alone cannot achieve this. Therefore, at the second level of the tree expansion, we introduce the concept of platforms. For each scenario, we generated multiple platforms with distinct characteristics. For example, in the video entertainment scenario, platforms such as YouTube and TikTok were included, where YouTube focuses on long-form videos and TikTok emphasizes short, lifestyle-oriented clips. This enables us to obtain multiple tools with functionally interchangeable capabilities.

\subsection{User Profile Construction}

% To introduce the concept of personalization, we need to construct various user profiles.
% When constructing the user profile, we face several challenges: first, determining which features should be included in the user profile and how these features should be linked to the tools to facilitate the subsequent construction of tool preference and incomplete query data; second, ensuring sufficient diversity between the profiles of different users, as without this, the model struggles to generalize to new users; and finally, ensuring that the user profile maintains a high degree of authenticity, including only fundamental, observable information (such as age, gender, etc.) and behavioral traits, without detailed psychological descriptions.

Personalization requires constructing diverse and realistic user profiles. This process involves three key challenges: (1) defining feature sets relevant to tool invocation, ensuring a structured linkage between user traits and tool selection; (2) maintaining sufficient diversity across profiles to enable generalization to unseen users; and (3) ensuring that profiles contain only observable basic and behavioral information, without incorporating detailed psychological attributes.

\textbf{Bottom-up Feature Tree Construction.} 
To systematically define user profile features, we adopt a tool-driven hierarchical clustering approach. We construct a feature tree, where platform characteristics and tool parameters serve as leaf nodes. Using advanced LLM-based clustering, we recursively merge semantically related parameters, summarizing them into higher-level features. This process continues until the number of parent nodes at each level falls within a predefined threshold.
Notably, we categorize features during initial clustering: explicit basic features (e.g., age, gender) are directly observable, while implicit preferences (e.g., shopping preferences) remain latent and are used in subsequent user behavior generation. 
% This structured organization allows for efficient profile synthesis while preserving realism in user modeling.

% To address the first challenge, we employ a tool-driven tree-based approach to constructing profile features. Specifically, we construct a user feature tree from the bottom up, with platform characteristics and tool parameters as leaf nodes. These parameters are then clustered, and an advanced LLM summarizes the descriptions within each cluster to serve as parent nodes. This process is recursively applied upwards until the number of parent nodes at any given layer falls within a specified threshold. Notably, during the initial clustering, we explicitly separate the features into basic user information (e.g., age, gender) and implicit features (e.g., shopping preferences, taste preferences), with the latter treated as the user's internal and hidden profile, not visible to the model. These implicit features serve as the basis for user simulation, which is then used to construct the user's observable behavioral features.

\textbf{Top-down Characteristic Assignment.} Once the user feature tree is constructed, we encounter the second issue: how to diversify the assignment of values to these features to generate distinct user profiles. When using an advanced LLM to assign $N$ different user features, two options typically arise: one is to assign all features for a single user at a time and repeat this process $N$ times; the other is to assign all features for $N$ users in one pass. The first method incurs higher inference costs and makes it challenging to avoid repetition across multiple generations, while the second is constrained by the model's context length limitation, especially when N or the number of features is large. Therefore, we adopt a top-down hierarchical assignment based on the tree structure. Specifically, for nodes at the $l$-th layer, we assign $k_l$ different values simultaneously, and for the $(l+1)$-th layer nodes, the model generates $k_{l+1}$ different values for each parent node's feature value. Thus, for a user feature tree with depth $L$, we can ultimately obtain $N = \prod_{l=0}^{L} k_l$ distinct user profiles. It's important to note that each time the LLM generates $k_l$, this number can be much smaller than $N$, allowing the LLM to generate diverse features in one pass.

\textbf{User Behavior Generation.} 
Once user profiles are assigned, they include both explicit basic features (e.g., occupation, gender, location) and implicit preferences (e.g., price sensitivity, product affinity). However, in real-world scenarios, user preferences are typically inferred through behavioral patterns rather than explicitly stated.
To simulate authentic behavioral traits, we employ an LLM-based role-playing approach, where the model generates user actions on various platforms based on their profile and platform characteristics. For instance, given a user's preference for budget-conscious shopping, the model may generate interactions such as "searches for hiking backpacks on Amazon" or "purchases coffee from Walmart for \$30."
While implicit preferences remain unobservable to the model during task execution, they are embedded in prompts when generating tool invocation solutions, ensuring accurate and contextually appropriate tool selection.

% After this assignment, we obtain the basic information features (such as occupation, gender, and address) and implicit features (such as price sensitivity, product preferences, and dietary preferences) for N users. However, in practical applications, psychological traits are often not directly expressed by users. Instead, these traits need to be inferred from user behavior. To ensure data authenticity, we use the generated features to conduct user simulation, thereby constructing explicit behavioral features. Specifically, given the user's basic and implicit features, along with the scenario, the platform in the scenario, and the platform's characteristics, we leverage the role-playing capabilities of the LLM to generate relevant behavioral descriptions within the context of the scenario. For example, "search for a hiking backpack on Amazon" or "purchase coffee from Walmart for \$30." At this point, the user profile construction is complete, primarily encompassing the user's basic information and behavioral features. While the implicit features remain hidden from the model, they are incorporated into the prompt during subsequent solution construction to ensure accuracy in the answers.

\subsection{Query and Solution Generation}

For generating query-solution pairs, we adopt a multi-agent collaborative approach, involving two agents: the user agent and the assistant agent. The user agent generates queries by role-playing based on the user profile, while the assistant agent generates tool invocation solutions. The user agent's role information includes both basic and implicit features, as these provide a more accurate user representation than explicit behavioral features.

Given that a user's platform preferences may vary across queries, we explicitly incorporate platform information into the user agent's prompt. This enables the agent to generate queries aligned with the user's platform preferences. Additionally, we instruct the user agent to avoid revealing profile information in the queries, ensuring the generation of profile-dependent queries as well.

To ensure the correctness of tool invocations, we employ a two-tier verification strategy: rule-based validation and model-based verification. Rule-based validation checks the format of tool invocations to prevent issues such as unresolvable results or hallucinated tools and parameters. Model-based verification inputs the user profile, query, and solution triples into the LLM to verify parameter correctness, detect hallucinations, and assess whether the solution effectively resolves the query. Furthermore, to ensure evaluation accuracy, we manually inspect the correctness of tool invocation parameters. These parameters are annotated as profile-related or query-related, indicating whether they originate from the user profile or the query, facilitating more precise error feedback during evaluation.

\section{Experiments}

\begin{table}[t]
    \centering
    \small
    \caption{Statistics of our synthesized dataset. The samples in the test set are verified by human annotators. Trained and untrained represent the user profiles present and absent in the training set, respectively.}
    \setlength{\tabcolsep}{2pt} % 设置列间隔为 10pt
    \begin{tabular}{lrrrrr}
        \toprule
        \textbf{Dataset} & \textbf{\#Scenario} & \textbf{\#Platform} & \textbf{\#API} & \textbf{\#User} & \textbf{\#Query} \\ \midrule
        \textbf{Train} & 5 & 15 & 360 & 74 & 7,096\\
        \midrule
        \textbf{Test(PTBench)} &  5 & 15 & 360 & 80 & 1,083\\
        --Trained &  5 & 15 & 360 & 74 & 474\\
        --Untrained &  5 & 15 & 360 & 6 & 609\\
        \midrule
        \textbf{Total} & 5 & 15 & 360 & 80 & 8,197\\
        \bottomrule
    \end{tabular}
    \label{tab:stat}
\end{table}

\subsection{Experimental Settings}
% experimental settings
% 训练数据描述：数据多少，#tool，user。。。
% evaluation 策略：指标
% baseline：和谁比，\footnote \cite
% IMPLEMENTATION DETAILS: 训了xx模型，用了什么技术lora，基本超参数，机器

\noindent\textbf{Dataset Details}. We leverage GPT-4-turbo to synthesize the personalized tool invocation dataset via our framework. The overall dataset consists of a total of 80 users and 7,096 queries under 5 scenarios, including shopping, takeout, entertainment, work, and travel. Under each scenario, there are 3 platforms and 24 APIs in each platform as tools. The data set is divided into training and test sets, randomly selecting all queries of 6 users and about 6\% queries of another 74 users to form the test set PTBench. The 6 users are not visible in the training process, termed as untrained. To ensure the quality of the test set, we manually verify each sample. The statistics are illustrated in Table~\ref{tab:stat}.

\begin{table*}[t]
    \centering
    \small
    \caption{Comparison with baseline models on PTBench. \textbf{Bold} and \underline{underline} represent the best and the 2nd best results. \textbf{Preference} denotes the ability of tool preference. \textit{T-*} denotes the ability of filling correctness * in tool invocation. \textbf{DS-R1-Dis} is the abbreviation of DeepSeek-R1-Distill. All the results are accuracy.}
    \label{tab:overall}
    \setlength{\tabcolsep}{2pt}{
    \begin{tabular}{c|l|cccccccccc}
    \toprule
        \multirow{2}{*}{\textbf{Type}} & \multirow{2}{*}{\textbf{Model}} &  \multirow{2}{*}{\textbf{Format}} & \textbf{Preference} & \multicolumn{2}{c}{\textbf{Param Value}} & \multicolumn{3}{c}{\textbf{Tool Invocation}} & \multicolumn{3}{c}{\textbf{Overall}} \\
        \cmidrule(lr){4-4} \cmidrule(lr){5-6} \cmidrule(lr){7-9} \cmidrule(lr){10-12}
         &  & & \textit{Platform} & \textit{Query} & \textit{Profile} & \textit{T-name} & \textit{T-param} & \textit{T-value} & \textit{Trained} & \textit{Untrained} & \textit{Overall} \\  \midrule
        \multirow{6}{*}{API} & \textbf{GPT-4-turbo} & \textbf{0.9778} & 0.5484 & \textbf{0.8123} & \underline{0.6832} & \underline{0.9178} & \underline{0.7709} & \textbf{0.3518} & \underline{0.1834} & \underline{0.1856} & \underline{0.1847}\\ 
        &\textbf{GPT-4o} & 0.9012 & 0.4484 & 0.7144 & 0.6104 & 0.8283 & 0.6991 & 0.2869 & 0.1350 & 0.1708& 0.1551\\ 
        &\textbf{Deepseek-v3} & 0.9095 & 0.5280 & 0.7309 & 0.6416 & 0.8460 & 0.7530 & 0.3085 & 0.1708 & 0.1757 & 0.1736\\ 
        &\textbf{Deepseek-r1} & 0.8199 & 0.4819 & 0.6304 & 0.5806 & 0.7376 & 0.6294 & 0.2624 & 0.1477 & 0.1494 & 0.1486\\ 
        &\textbf{Qwen-max} & 0.7692 & 0.4946 & 0.6094 & 0.5440 & 0.7091 & 0.5843 & 0.2348 & 0.1456 & 0.1707 & 0.1597\\ 
        &\textbf{Claude-3.5-sonnet}  & \underline{0.9686} & \underline{0.5826} & 0.7824 & 0.6504 & 0.7110 & 0.6445 & 0.2326 & 0.1329 & 0.1395 & 0.1367 \\  \midrule
        \multirow{10}{*}{OSS} & \textbf{DS-R1-Dis-Llama-8B} & 0.6427 & 0.3019 & 0.3823 & 0.3012  & 0.5080 & 0.3802 & 0.0981 & 0.0485 & 0.0394 & 0.0434 \\ 
        &\textbf{DS-R1-Dis-Qwen-7B} & 0.6095 & 0.1469 & 0.2341 & 0.1039 & 0.3656 & 0.2113 & 0.0221 & 0.0042 & 0.0066 & 0.0055 \\ 
        &\textbf{Qwen2.5-7B-Instruct} & 0.7858 & 0.3795 & 0.6132 & 0.4165 & 0.6833 & 0.5430 & 0.1837 & 0.0717 & 0.0755 &  0.0738 \\ 
        &\textbf{Llama-3.1-8B-Instruct} & 0.8865 & 0.4053 & 0.6648 & 0.5141 & 0.7997 & 0.6252 & 0.2133 & 0.0929 & 0.0985 & 0.0960 \\ 
        &\textbf{Mistral-7B-Instruct-v0.3} & 0.8587 & 0.3903 & 0.5598 & 0.3723 & 0.6612 & 0.3572 & 0.1450 & 0.0674 & 0.0559 & 0.0609\\ 
        &\textbf{Hammer2.1-7b} & 0.9649 & 0.3638 & 0.7296 & 0.5259 & 0.8402 & 0.6316 & 0.2262 & 0.0739 & 0.0689 & 0.0711 \\ 
        &\textbf{ToolACE-8B} & 0.4035 & 0.1681 & 0.3289 & 0.2049 & 0.3887 & 0.2631 & 0.0906 & 0.0338 & 0.0378 & 0.0360\\ 
        &\textbf{Watt-tool-8B} & 0.3749  & 0.2281 & 0.2716 & 0.1990 & 0.3408 & 0.2218 & 0.0826 & 0.0591 & 0.0411 & 0.0489 \\ 
        &\textbf{xLAM-7b-r} & 0.9529 & 0.3285 & 0.6794 & 0.4968 & 0.8688 & 0.5934 & 0.2217 & 0.0696 & 0.0771 & 0.0738 \\  \midrule
        &\textbf{Ours} & 0.9575 & \textbf{0.7374} & \underline{0.7933} & \textbf{0.7341} & \textbf{0.9242} & \textbf{0.8290} & \underline{0.3417} & \textbf{0.2701} & \textbf{0.2660} & \textbf{0.2678}\\ 
        \bottomrule
    \end{tabular}}
\end{table*}

\noindent\textbf{Evaluation}.
We first evaluate the format accuracy by checking if the model's output can give formatted output, verifying the instruction following ability.
The solution of each sample comprises two major parts: the platform and the tool invocation. The models are required to select the correct user-preferred platform and then generate suitable tool invocations. The platform accuracy demonstrates the ability of tool preference understanding. The tool invocation consists of three parts: tool name, parameters, and parameter values, where the parameter values comprise query-related and profile-related parameters. The profile-related parameters require the model to infer from the user profile, evaluating the ability to handle profile-dependent query. We calculate the accuracy of the function name, function parameter, and function value, respectively. The calculations of accuracy are detailed in Appendix~\ref{sec:appd_metric}.

% We divide the answer into two parts: platform and functions. For platform part, we directly calculate the accuracy. For functions part, we divide them into three phases: function names, function parameters and function values. We calculate the accuracy of the phases progressively. Then we synthesize the accuracy of platform and the functions into the overall accuracy.

\noindent\textbf{Baselines}. We compare the latest open-source models and API-based models, as well as fine-tuned tool-calling models. Open-source models include DeepSeek-R1-Distill-Llama-8B\cite{deepseekai2025deepseekr1incentivizingreasoningcapability}, DeepSeek-R1-Distill-Qwen-7B\cite{deepseekai2025deepseekr1incentivizingreasoningcapability}, Qwen2.5-7B-Instruct\cite{qwen2, qwen2.5}, Llama-3.1-8B-Instruct~\cite{llama3modelcard} and Mistral-7B-Instruct-v0.3\cite{jiang2023mistral7b}. API-based models include GPT-4-turbo\footnote{https://chatgpt.com\label{gpt}}, GPT-4o\textsuperscript{\ref{gpt}}, Deepseek-v3\cite{deepseekai2024deepseekv3technicalreport}, Deepseek-r1\cite{deepseekai2025deepseekr1incentivizingreasoningcapability}, Qwen-max\cite{qwen2.5} and Claude-3.5-sonnet\footnote{https://www.anthropic.com}. Models fine-tuned for tool-calling include Hammer2.1-7b\cite{lin2024hammer}, ToolACE-8B\cite{liu2025toolace}, watt-tool-8B\footnote{https://ollama.com\label{watt}} and xLAM-7b-r\cite{zhang2024xlam, liu2024apigen, zhang2024agentohana}.

 \noindent\textbf{Implementation Details}. To validate the effectiveness of our model, we conducted various experiments by training LLMs with the synthesized dataset. We train the open-source LLM, Qwen2.5-7B-Instruct\cite{qwen2, qwen2.5}, in the supervised fine-tuning (SFT) manner. Due to limited resources, we adopt the parameter-efficient LoRA\cite{hu2022lora} training strategy to fine-tune the model. As for the hyper-parameters setting, we set the rank as 8, alpha as 16 learning rate as $10^{-4}$, LR scheduler as cosine, WarmUp Ratio as 0.1 and epoch as 1 for all modules in the model. 
 % More detailed training settings are shown in Table~\ref{tab:Hyper-parameters}.

% \iffalse
% \begin{table*}[t]
%     \centering
%     \small
%     \caption{Training Hyper-parameters}
%     \label{tab:Hyper-parameters}
%     \setlength{\tabcolsep}{3mm}{
%     \begin{tabular}{ccccccc}
%     \toprule
%     Learning Rate & WarmUp Ratio &  LR Scheduler & Batch Size & Epochs & LoRA rank & LoRA alpha \\ \midrule
%     $10^{-4}$ & 0.1 & cosine & 1 & 1 & 8 & 16 \\
%     \bottomrule
%     \end{tabular}}
% \end{table*}
% \fi

\subsection{Main Results}
% 简单的设定。TRAINED USER展示在xxx，
The overall results are illustrated in Table~\ref{tab:overall}. The detailed results of trained and untrained users are presented in Appendix~\ref{sec:detailed_results}. We have the following findings according to the results:

\textit{Finding 1:}
API-based large models significantly outperform smaller OSS models across various dimensions, including format compliance, tool preference capabilities, and tool invocation abilities. This aligns with the findings of most benchmarks, primarily attributed to the enhanced capabilities enabled by the larger scale of model parameters.

\textit{Finding 2:} 
Most models fall short on the tool preference task, demonstrating low platform accuracy, including the state-of-the-art advanced model GPT-4-turbo. This phenomenon indicates that most LLMs fail to select suitable tools according to the user profile.
Our model outperforms nearly all models in all aspects by a considerable improvement, presenting the necessity of personalized tool-invocation enhancement.

\textit{Finding 3:}
Our model demonstrates a significant improvement in its performance across various tasks on PTBench. Notably, the enhancement in the Tool Preference task is particularly pronounced when compared to the pre-trained Qwen2.5-7B-Instruct model. This also indicates that, even without additional manual verification of the training data, the model achieves a high accuracy, demonstrating the effectiveness of the proposed synthesis framework. Additionally, our model shows a significant improvement on untrained users, presenting the generalization of the model.

\textit{Finding 4:}
All models exhibit lower accuracy on profile-dependent parameter values compared to query-dependent parameters, indicating that inferring parameters from the profile presents a greater challenge. While our trained model does not surpass GPT-4-turbo in accuracy on query-dependent parameters, it outperforms larger models on profile-dependent parameters. Furthermore, the improvement over the pre-trained Qwen2.5-7B-Instruct model is more substantial, demonstrating the effectiveness of our data generation framework in handling the query-dependent query tasks.

\subsection{Ablation Study}

% 结果
% 分析
To investigate the importance of various parts in our synthesized user profile, we conduct the ablation study on the user profile, including 4 variants on the user profile:
\begin{itemize}[leftmargin=*]
    \item \textbf{All.} All information in the user profile is used, including basic features and behavioral history.
    \item \textbf{All w/o Basic.} Basic features are omitted.
    \item \textbf{All w/o History.} The behavioral history is given.
    \item \textbf{All w/o Basic\&History.} Both basic features and behavioral history are omitted.
\end{itemize}

First, We use the four dataset variants to train and then evaluate the model with the consistent input. The results are reported in Table~\ref{tab:ablation_consistent}. From the result, we can observe that the existence of user history and basic features hold contributions to the overall performance of the model to an extent.

\begin{figure}[thb]
    \centering
    \begin{subfigure}{0.45\textwidth}
        \centering
        \includegraphics[width=\textwidth]{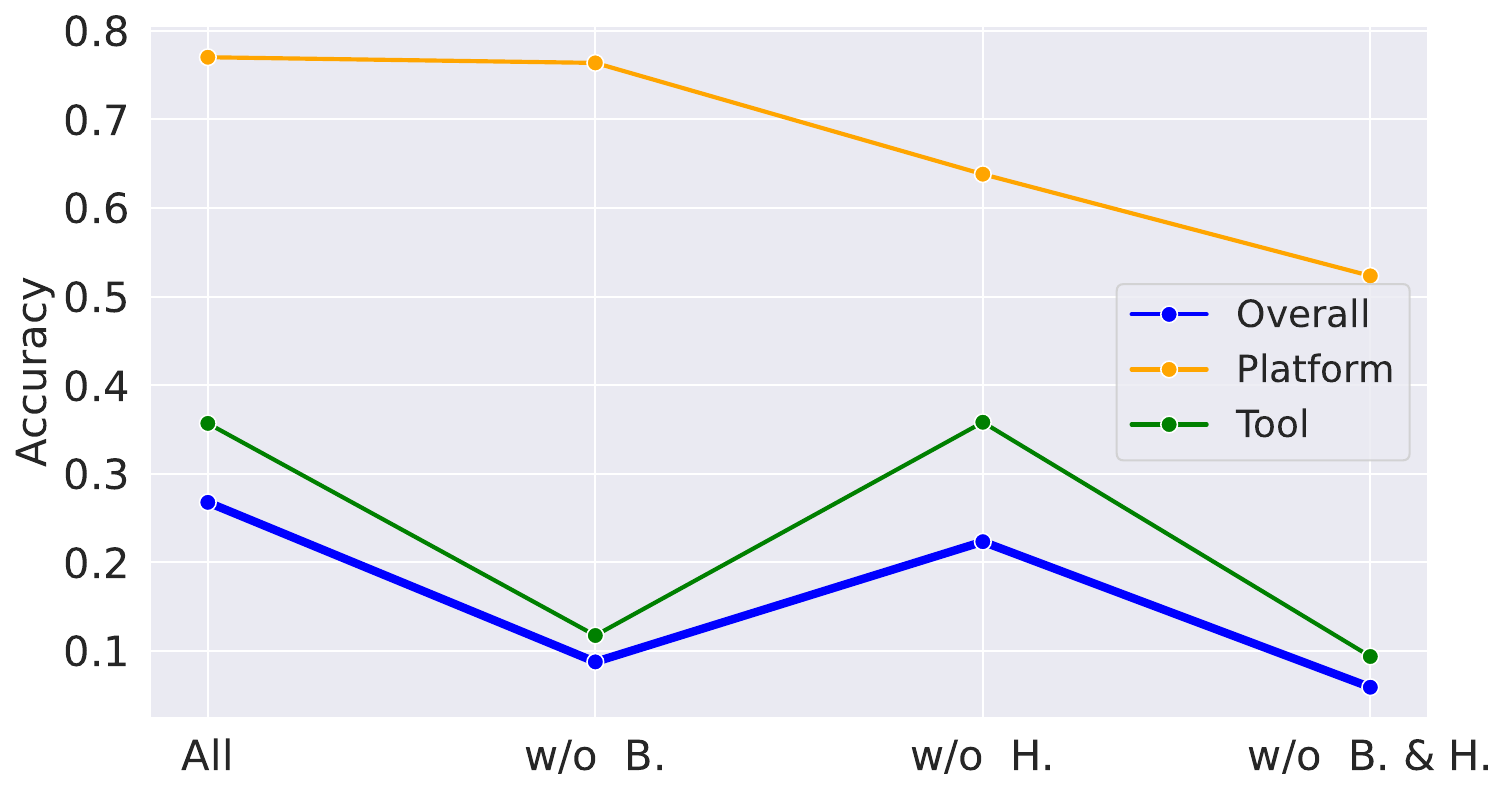}
        \caption{User profile ablation in evaluation.}
        \label{fig:test_ablation}
    \end{subfigure}
    \begin{subfigure}{0.45\textwidth}
        \centering
        \includegraphics[width=\textwidth]{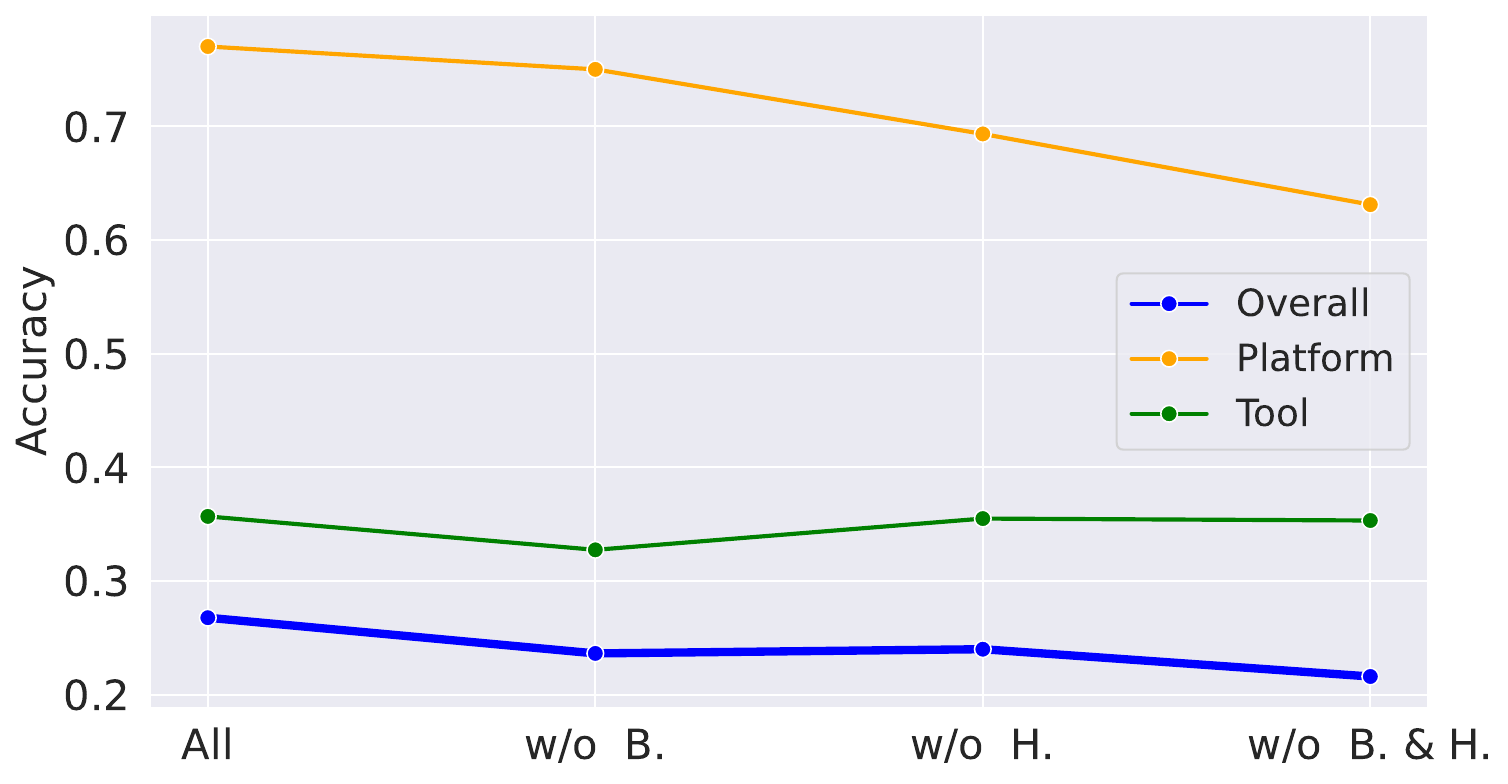}
        \caption{User profile ablation in training.}
        \label{fig:train_ablation}
    \end{subfigure}
    \caption{Ablation study on user profile in evaluation and training, respectively. We abbreviate "Basic" and "History" as "B." and "H." respectively, and omit the preceding "All."}
\end{figure}

Additionally, we conduct experiments under two settings: (1) train the model with the All variant and evaluate the model with the four variants, illustrated in Figure~\ref{fig:test_ablation}; (2) train the model with the four variants and evaluate the models with the All variant, illustrated in Figure~\ref{fig:train_ablation}.  The results exhibit that the model shows poor performance in the tool preference task when lacking user history information in training or evaluation. 
On the other hand, the accuracy of tool invocation suffers when basic features are absent, led by the challenging profile-dependent query task.
% while basic features improve model's tool-invocation accuracy, user history tend to promote model's platform accuracy, presenting that the two components contribute to different aspect of the model's capability. 

% In the second and third study, we set the train set and test set as all-known dataset respectively, and switch the other dataset in order to investigate the ablation in test set and train set. The results are shown in Figure~\ref{fig:test_ablation} and Figure~\ref{fig:train_ablation}.

% Taken together, the results of the three studies provide strong evidence that each component of the dataset plays a crucial role in achieving optimal performance. Through comprehensive analysis and validation, it is clear that all components are equally essential for maximizing the model’s effectiveness. This underscores the importance of considering the dataset holistically, as omitting any of its key parts could significantly diminish the overall results.

\begin{table}[t]
    \centering
    \small
    \caption{Ablation of user profile on PTBench. The models are trained with various variants. The input in evaluation remains consistent with the training input.}
    \label{tab:ablation_consistent}
    \setlength{\tabcolsep}{1mm}{
    \begin{tabular}{lcccc}
    \toprule
         \textbf{Data} & \textbf{Untrained} & \textbf{Trained} & \textbf{Overall}\\ \midrule
        \textbf{All} & \textbf{0.2660}  & \textbf{0.2701} &  \textbf{0.2678}\\
        \textbf{All w/o Basic} & 0.0969 & 0.2426 & 0.1606\\
        \textbf{All w/o History} & 0.2463 & 0.2531 & 0.2493\\
        \textbf{All w/o Basic\&History} & 0.0591 & 0.0781 & 0.0674\\
        \bottomrule
    \end{tabular}}
\end{table}

\subsection{Error Analysis}
% TO BE CONTINUE
In the intention of gaining deeper insights into the function errors made by the models during the evaluation, we conduct investigations on the errors. We specifically choose our model, GPT-4-turbo and Qwen2.5-7B-Instruct to continue our investigation. We only analyze solutions with the correct format.

\begin{figure*}[t]
    \centering
    \includegraphics[width=1\linewidth]{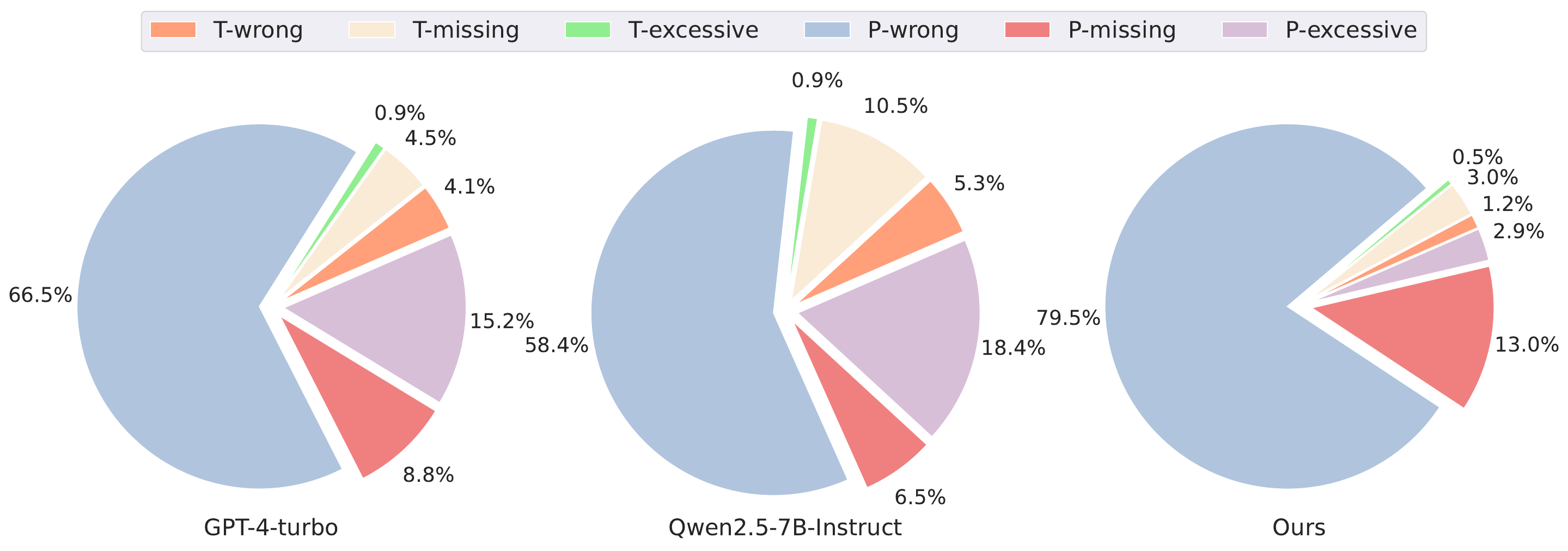}
    \caption{Error Analysis on PTBench. T-wrong, T-missing, and T-excessive represent wrong tools, missing tools and excessive tools, respectively. P-missing, P-excessive and P-error represent missing parameters, excessive parameters and wrong parameters, respectively.}
    \label{fig:overall_error_analysis}
\end{figure*}

We analyze the function errors generally and divide them into 6 categories: wrong tools, missing tools, excessive tools, missing parameters, excessive parameters, and wrong parameters. The results are shown in Figure~\ref{fig:overall_error_analysis}. From the pie chart, it is evident that filling the correct parameters is more challenging than the selection of the correct tools. After training with our synthesized data, the model is more familiar with the candidate tools, demonstrating less error percentage in tool selection.
% most of the errors are caused by parameter-erro because of the difficulty of filling in the parameter values. Moreover, the observation indicates that our model tend to miss some parameters while the other models tend to include excessive parameters. 

% In the second study, we look deeper into the main category of errors, parameter-wrong. We divide the parameters into two categories: query and profile, indicating where the parameter value can be found. We investigate the accuracy of the two kinds of parameters and the results are presented in Figure~\ref{fig:param_error_analysis}. The figures demonstrate that although our model do not perform at the level of GPT-4-turbo\textsuperscript{\ref{gpt}} on query parameters, our model excels on profile parameters and outperforms the base model, Qwen2.5-7B-Instruct, by a large margin in both categories, showing the our model's outstanding capability of interpreting information from the long context.

% \begin{figure}[htb]
%     \centering
%     \includegraphics[width=1\linewidth]{pics/param_error_bar.pdf}
%     \caption{Parameter Error Analysis}
%     \label{fig:param_error_analysis}
% \end{figure}

\subsection{Further Analysis}

\begin{figure}[t]
    \centering
    \includegraphics[width=1\linewidth]{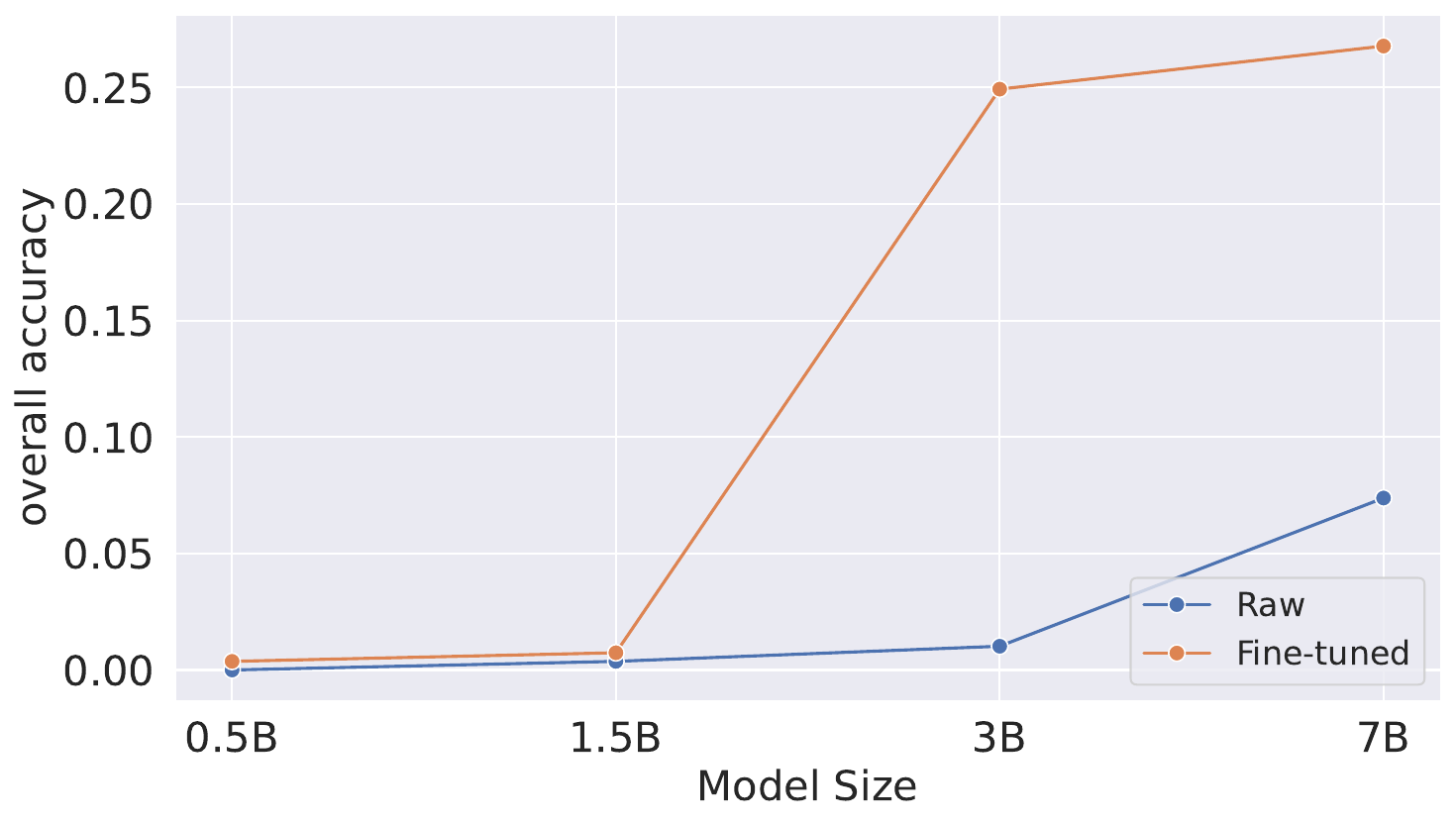}    
    \caption{Study of model scaling. The base models are Qwen2.5-series.}
    \label{fig:model_scaling}
\end{figure}

\textbf{Model Scaling.} 
% 设定：为了xxx，我们选择了xxx进行训练，训练结果展示在图xx中。从图中，我们发现xxx
For the purpose of analyzing the influence of model size on the performance of our trained model, we utilize models with different sizes in the Qwen2.5 series, including 7B, 3B, 1.5B and 0.5B. The results are shown in Figure~\ref{fig:model_scaling}. We can observe that the 1.5B and 0.5B model only show slight improvement from the training, while 3B and 7B model gain substantial improvement from the training. This demonstrate that the personalized tool invocation is a high-level capability of LLMs, requiring a certain scale of parameters.
 
\begin{figure}[thb]
    \centering
    \includegraphics[width=0.8\linewidth]{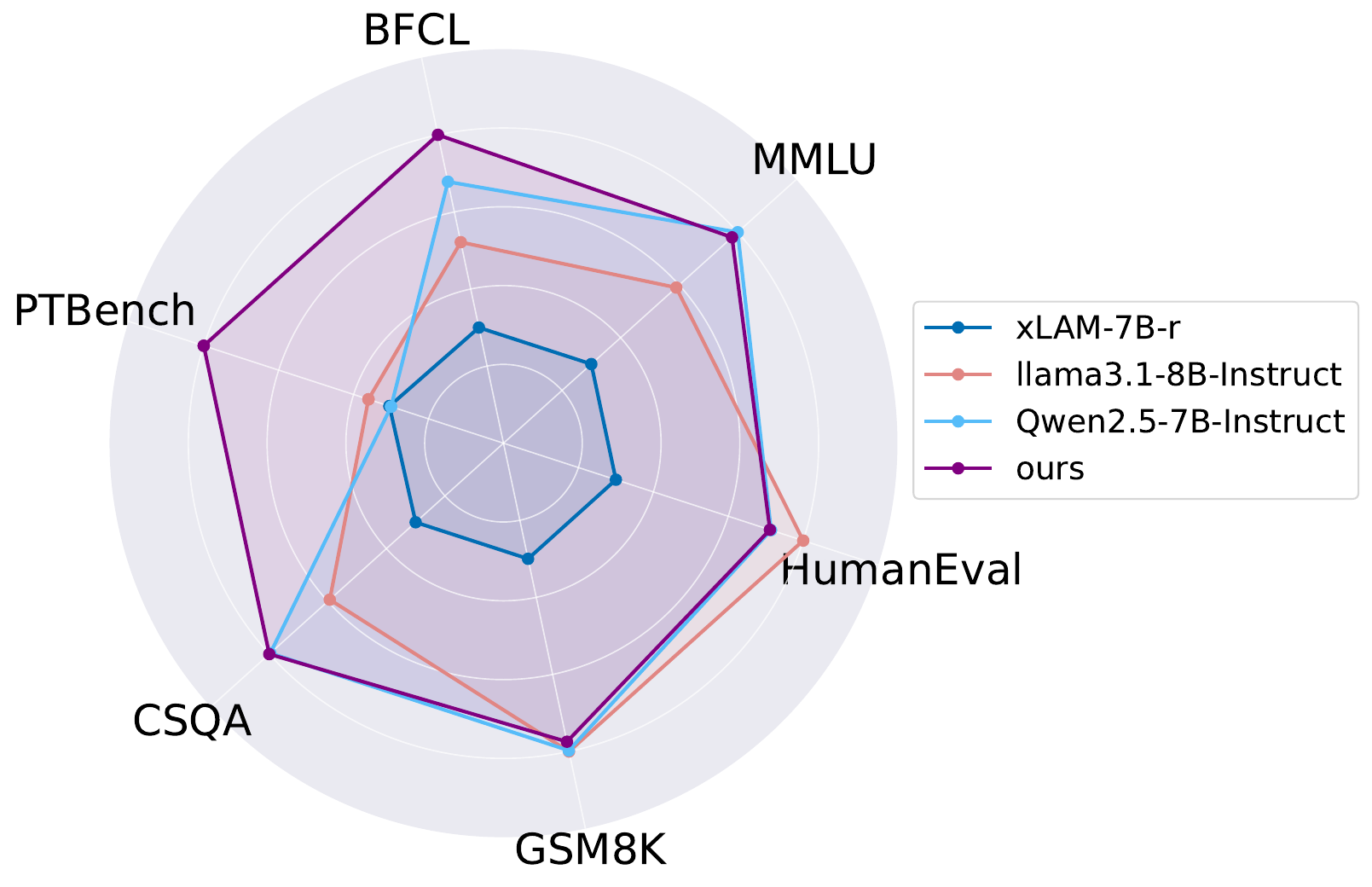}
    \caption{General Capabilities Analysis. Our model is fine-tuned from Qwen2.5-7B-Instruct.}
    \label{fig:general_capability_analysis}
\end{figure}

\noindent\textbf{General Capabilities.}
% 为了xxx（验证我们的训练数据/目标是否影响模型的其他能力），我们xxxx。
In order to validate that our synthesized data does not introduce negative effects on the model's general capabilities, we employ a diverse set of benchmarks to assess the performance from different perspectives, including general ability(MMLU\cite{hendrycks2021ethics, hendryckstest2021}), coding(HumanEval\cite{chen2021codex}), math(GSM8K\cite{cobbe2021gsm8k}), reasoning(CommonSenceQA\cite{talmor-etal-2019-commonsenseqa}) and basic function calling(tool-invocation) ability (BFCL non-live\cite{berkeley-function-calling-leaderboard}). xLAM-7B-r, LLaMA-3-8B-Instruct, Raw Qwen2.5-7B-Instruct serve as baselines. The results are shown in Figure~\ref{fig:general_capability_analysis}. From the figure, it is evident that there is no significance deterioration on abilities of our model compared to the raw model Qwen2.5-7B-Instruct. Nonetheless, our model gains a notable improvement on BFCL non-live,  These findings suggest that our approach effectively enhances personalized functional calling capabilities without compromising the underlying LLM’s other abilities.

\section{Conclusion}
% 在这个工作中，我们创新性的提出了个性化工具调用的概念，其中包含两个主要任务：tool preference和 profile-dependent query。其中要求到模型对用户profile的理解能力，根据用户历史行为选择用户偏好工具的能力，从用户信息中提取工具参数的能力。为了提升和测评模型personalized tool-invocation的能力，我们提出了一个数据合成框架，并对划分出部分数据进行人工检查后作为benchmark——PTBench。Extensive的实验评估了现有模型的个性化工具调用的能力，并验证了我们合成的数据的有效性和对模型其他能力的无害性。

In this work, we introduce the concept of personalized tool invocation, which encompasses two primary tasks: tool preference and profile-dependent queries. These tasks require the model’s ability to understand the user's profile, select preferred tools based on historical behavior, and extract tool parameters from user information. To enhance and evaluate the model's personalized tool invocation capabilities, we propose a data synthesis framework and create a benchmark, PTBench, by manually inspecting a subset of the generated data. Extensive experimental evaluations assess the personalized tool invocation abilities of existing models, confirming the effectiveness of our synthesized data and its harmlessness to other model capabilities.
\newpage
\section*{Limitations}
We conclude the limitations of this work as follows:

First, the current coverage of scenarios is limited, as we primarily focus on the five most commonly encountered scenarios in daily life. However, this does not encompass the full spectrum of everyday needs. We plan to expand the range of scenarios covered by our tools in future work.

Second, personalized tool invocation is a crucial ability for LLMs in daily life. While we have proposed two key tasks in this work, they do not fully capture the entire scope of personalized tool invocation. One direction for future development is to introduce additional tasks that better address the diverse requirements of personalized tool invocation.

\bibliography{mybib,anthology}

\appendix

\section{Experiments}

\subsection{Evaluation Metrics}\label{sec:appd_metric}
The calculation of various metrics in PTBench are formulated as follows:
\begin{itemize}[leftmargin=*]
    \item \textbf{Format Accuracy} indicates the instruction-following ability.
    \begin{small}
    \begin{equation}
        {format\_acc}=\frac{\#parsable\,samples}{\#total}
    \end{equation}
    \end{small}
    \item \textbf{Platform Accuracy} indicates the tool preference recognition ability. 
    \begin{small}
    \begin{equation}
        {platform\_acc}=\frac{\#correct\,platform\,samples}{\#total}
    \end{equation}
    \end{small}
    \item \textbf{Query-related Parameter-Value Accuracy} indicates the ability to extract values from query.
    \begin{small}
    \begin{equation}
        {query\_param\_acc}=\frac{\#correct\,query\,params}{\#total\,query\,params}
    \end{equation}
    \end{small}
    \item \textbf{Profile-related Parameter-Value Accuracy} indicates the ability to extract values from profile.
    \begin{small}
    \begin{equation}
        {profile\_param\_acc}=\frac{\#correct\,profile\,params}{\#total\,profile\,params}
    \end{equation}
    \end{small}
    \item \textbf{Tool Name Accuracy} indicates the tool selection ability.
    \begin{small}
    \begin{equation}
        {tool\_name\_acc}=\frac{\#correct\,name\,samples}{\#total}
    \end{equation}
    \end{small}
    \item \textbf{Tool Parameter Accuracy} indicates the tool comprehension ability.
    \begin{small}
    \begin{equation}
        {tool\_param\_acc}=\frac{\#correct\,param\,samples}{\#total}
    \end{equation}
    \end{small}
    \item \textbf{Tool Parameter-Value Accuracy} indicate the value extraction on context ability.
    \begin{small}
    \begin{equation}
        {tool\_value\_acc}=\frac{\#correct\,value\,samples}{\#total}
    \end{equation}
    \end{small}
    \item \textbf{Overall Accuracy on Trained Users} indicate the personalized tool ability on trained users.
    \begin{small}
    \begin{equation}
        {trained\_overall\_acc}=\frac{\#correct\,trained\,samples}{\#trained\,total}
    \end{equation}
    \end{small}
    \item \textbf{Overall Accuracy on Untrained Users} indicate the personalized tool selection ability on trained users.
    \begin{small}
    \begin{equation}
        {untrained\_overall\_acc}=\frac{\#correct\,untrained\,samples}{\#untrained\,total}
    \end{equation}
    \end{small}
    \item \textbf{Overall Accuracy} indicate the overall personalized tool selection ability.
    \begin{small}
    \begin{equation}
        {overall\_acc}=\frac{\#correct\,samples}{\#total}
    \end{equation}
    \end{small}
\end{itemize}

\subsection{Detailed Results}\label{sec:detailed_results}
The detailed results of the trained and untrained subset on PTBench are illustrated in Table~\ref{tab:overall_trained} and Table~\ref{tab:overall_untrained}, respectively.

\section{Examples}
To enhance the understanding of the proposed personalized tool invocation, we illustrate an example in Figure~\ref{fig:prompt_generation}.

\begin{table*}[t]
    \centering
    \small
    \caption{Comparison with baseline models on trained users in PTBench. \textbf{Bold} and \underline{underline} represent the best and the 2nd best results.}
    \label{tab:overall_trained}
    \setlength{\tabcolsep}{2pt}{
    \begin{tabular}{c|l|cccccccc}
    \toprule
        \multirow{2}{*}{\textbf{Type}} & \multirow{2}{*}{\textbf{Model}} &  \multirow{2}{*}{\textbf{Format}} & \textbf{Preference} & \multicolumn{2}{c}{\textbf{Param Value}} & \multicolumn{3}{c}{\textbf{Tool Invocation}} & \multirow{2}{*}{\textbf{Overall}}\\
        \cmidrule(lr){4-4} \cmidrule(lr){5-6} \cmidrule(lr){7-9} \cmidrule(lr){10-10}
         &  & & \textit{Platform} & \textit{Query} & \textit{Profile} & \textit{T-name} & \textit{T-param} & \textit{T-value} & \\  \midrule
        \multirow{6}{*}{API} & \textbf{GPT-4-turbo} & \textbf{0.9831} & 0.5569 & \textbf{0.7927} & \underline{0.7080} & \underline{0.9325} &\underline{0.7869} &\textbf{0.3502} &\underline{0.1834} \\ 
        &\textbf{GPT-4o} & 0.8840 &0.4157 & 0.6520 & 0.6164 & 0.8143 &0.6941 &0.2637 &0.1350 \\ 
        &\textbf{Deepseek-v3} & 0.8903 &0.5043 & 0.6868 & 0.6508 & 0.8376 &0.7617 &0.3059 &0.1708 \\ 
        &\textbf{Deepseek-r1} &0.8376 &0.4958 & 0.6112 & 0.6317 & 0.7637 &0.6604 &0.2574 &0.1477 \\ 
        &\textbf{Qwen-max} & 0.6941 &0.4430 & 0.5083 &0.5162 & 0.6393 &0.5358 &0.2152 &0.1456 \\ 
        &\textbf{Claude-3.5-sonnet}  & \underline{0.9662} &\underline{0.5822} & 0.7519 & 0.6794 & 0.7152 &0.6498 &0.2236 &0.1329  \\  \midrule
        \multirow{10}{*}{OSS} & \textbf{DeepSeek-R1-Distill-Llama-8B} & 0.6203 &0.2891 & 0.3495 & 0.3111 & 0.4958 &0.3925 &0.1013 &0.0485  \\ 
        &\textbf{DeepSeek-R1-Distill-Qwen-7B} & 0.6013 &0.1519 & 0.2148 & 0.0954 & 0.3503 &0.1941 &0.0147 &0.0042  \\ 
        &\textbf{Qwen2.5-7B-Instruct} & 0.7827 &0.3882 & 0.5900 & 0.4447 & 0.6856 &0.5612 &0.1772 &0.0717  \\ 
        &\textbf{Llama-3.1-8B-Instruct} & 0.8819 &0.3797 & 0.6384 & 0.5439 & 0.8039 &0.6498 &0.2236 &0.0929  \\ 
        &\textbf{Mistral-7B-Instruct-v0.3} & 0.8713 &0.4198 & 0.5522 & 0.4113 & 0.6645 &0.3734 &0.1477 &0.0674 \\ 
        &\textbf{Hammer2.1-7b} & 0.9641 &0.3650 & 0.7126 & 0.5468 & 0.8439 &0.6582 &0.2257 &0.0739  \\ 
        &\textbf{ToolACE-8B} & 0.4114 &0.1709 & 0.3147 & 0.2061 & 0.3987 &0.2721 &0.0865 &0.0338 \\ 
        &\textbf{Watt-tool-8B} & 0.3966 & 0.2405 & 0.2708 & 0.2156 & 0.3586 &0.2510 &0.0992 &0.0591  \\ 
        &\textbf{xLAM-7b-r} & 0.9641 &0.3586 & 0.6732 & 0.5315 & 0.8881 &0.6329 &0.2194 &0.0696  \\  \midrule
        &\textbf{Ours} & 0.9662  &\textbf{0.7826} & \underline{0.7791} & \textbf{0.7653} & \textbf{0.9409} &\textbf{0.8628} &\underline{0.3333} & \textbf{0.2701} \\ 
        \bottomrule
    \end{tabular}}
\end{table*}

\begin{table*}[ht]
    \centering
    \small
    \caption{Comparison with baseline models on untrained users in PTBench. \textbf{Bold} and \underline{underline} represent the best and the 2nd best results.}
    \label{tab:overall_untrained}
    \setlength{\tabcolsep}{2pt}{
    \begin{tabular}{c|l|cccccccc}
    \toprule
        \multirow{2}{*}{\textbf{Type}} & \multirow{2}{*}{\textbf{Model}} &  \multirow{2}{*}{\textbf{Format}} & \textbf{Preference} & \multicolumn{2}{c}{\textbf{Param Value}} & \multicolumn{3}{c}{\textbf{Tool Invocation}} & \multirow{2}{*}{\textbf{Overall}}\\
        \cmidrule(lr){4-4} \cmidrule(lr){5-6} \cmidrule(lr){7-9} \cmidrule(lr){10-10}
         &  & & \textit{Platform} & \textit{Query} & \textit{Profile} & \textit{T-name} & \textit{T-param} & \textit{T-value} & \\  \midrule
        \multirow{6}{*}{API} & \textbf{GPT-4-turbo} & \textbf{0.9737} & 0.5419 & \textbf{0.8266} & \underline{0.6637} & \underline{0.9064} & \underline{0.7586} & \textbf{0.3531} & \underline{0.1856}  \\ 
        &\textbf{GPT-4o} & 0.9146 & 0.4746 & 0.7596 & 0.6057 & 0.8391 & 0.7028 & 0.3054 & 0.1708  \\ 
        &\textbf{Deepseek-v3} & 0.9245 & 0.5468 & 0.7629 & 0.6343 & 0.8522 & 0.7455 & 0.3104 & 0.1757  \\ 
        &\textbf{Deepseek-r1} & 0.8062 & 0.4712 & 0.6443 & 0.5403 & 0.7175 & 0.6059 & 0.2660 & 0.1494  \\ 
        &\textbf{Qwen-max} & 0.8276 & 0.5353 & 0.6828 & 0.5658 & 0.7635 & 0.6207 & 0.2496 & 0.1707  \\ 
        &\textbf{Claude-3.5-sonnet}  & \underline{0.9704} & \underline{0.5829} & \underline{0.8046} & 0.6275 & 0.7077 & 0.6404 & 0.2397 & 0.1395  \\  \midrule
        \multirow{10}{*}{OSS} & \textbf{DeepSeek-R1-Distill-Llama-8B} & 0.6601 & 0.3120 & 0.4061 & 0.2935 & 0.5173 & 0.3695 & 0.0953 & 0.0394   \\ 
        &\textbf{DeepSeek-R1-Distill-Qwen-7B} & 0.6158 & 0.1429 & 0.2481 & 0.1106 & 0.3777 & 0.2250 & 0.0279 & 0.0066   \\ 
        &\textbf{Qwen2.5-7B-Instruct} & 0.7882 & 0.3727 & 0.6301 & 0.3943 & 0.6815 & 0.5287 & 0.1889 & 0.0755   \\ 
        &\textbf{Llama-3.1-8B-Instruct} & 0.8900 & 0.4253 & 0.6839 & 0.4906 & 0.7964 & 0.6059 & 0.2052 & 0.0985   \\ 
        &\textbf{Mistral-7B-Instruct-v0.3} & 0.8489 & 0.3678 & 0.5653 & 0.3416 & 0.6584 & 0.3448 & 0.1429 & 0.0559  \\ 
        &\textbf{Hammer2.1-7b} & 0.9655 & 0.3629 & 0.7420 & 0.5094 & 0.8374 & 0.6109 & 0.2266 & 0.0689   \\ 
        &\textbf{ToolACE-8B} & 0.3974 & 0.1659 & 0.3392 & 0.2039 & 0.3810 & 0.2562 & 0.0936 & 0.0378  \\ 
        &\textbf{Watt-tool-8B} & 0.3580 & 0.2184 & 0.2722 & 0.1859 & 0.3268 & 0.2003 & 0.0706 & 0.0411   \\ 
        &\textbf{xLAM-7b-r} & 0.9442 & 0.3054 & 0.6839 & 0.4695 & 0.8538 & 0.5632 & 0.2233 & 0.0771   \\  \midrule
        &\textbf{Ours} & 0.9507 & \textbf{0.7028} & 0.8035 & \textbf{0.7096} & \textbf{0.9112} & \textbf{0.8030} & \underline{0.3481} & \textbf{0.2660}  \\ 
        \bottomrule
    \end{tabular}}
\end{table*}

\begin{figure*}[htb]
    \centering
    \scriptsize
\begin{mdframed}
\textsc{\textbf{[system]}}\\
You are given a user profile:
\begin{verbatim}
{
  "basic_features":{
        "username":"WineTraveler38",
        ...
  }
  "user_history":{
    "shopping":[{
            "platform":"MegaMart",
            "action":"Purchased a selection of premium imported wines"
        }
        ...
    ]
    ...
  }
}
\end{verbatim}
Here is some platforms under the scenario:
\begin{verbatim}
[{
        "name":"MegaMart",
        "profile":{
            "product range":"A wide-ranging selection, offering products from various categories."
            ...
        }
    }
    ...
]
\end{verbatim}
Here is some APIs under the platforms:
\begin{verbatim}
[{
        'type': 'function', 
        'function': {
            'name': 'registerUser', 
            'description': 'Registers a new user in the application.', 
            'parameters': {
                'type': 'object', 
                'properties': {
                    'username': {
                        'type': 'string', 
                        'description': \"User's chosen username.\"
                    },
                    ...
                }, 
                'required': ['username', 'password', 'email']
            }, 
            'response': {
                'type': 'object', 
                'properties': {
                    'success': {
                        'type': 'boolean', 
                        'description': 'Status of registration.'
                    },
                    ...
                }
            }
        }
    },
    ...    
]
\end{verbatim}
The user will give you a query. Based on the profile, try to solve the query by using the platforms and APIs. The platform you choose should fit the user profile or the needs of the user's query. All the necessary information are provided in the user profile. DO NOT ask the user for further information. You should respond in the format of \{platform:[func1(param1\_name = param1\_value, param2...), func2...]\} No other text MUST be included.\\

\textsc{\textbf{[user]}}\\
Could you please register an account for me using my username, password and email address, and setting my home location to my place of residence? I prefer not to receive any marketing emails.\\

\textsc{\textbf{[Assistant]}}
\color{blue}
\begin{verbatim}
{
  MegaMart:[
    registerUser(
      username='WineTraveler38', password='strongpassword123!',
      email='jeanlucbordeaux@email.com', preferredLanguage='French',
      marketingConsent=False, homeLocation='Paris, France'
    )
  ]
}
\end{verbatim}
\end{mdframed}
    \caption{Illustration of personalized tool invocation task. The blue part represents the expected output.}
    \label{fig:prompt_generation}
\end{figure*}

\end{document}